CLCLSA: Cross-omics Linked embedding with Contrastive Learning and Self Attention for multi-omics integration with incomplete multi-omics data

Chen Zhao[1], Anqi Liu[2], Xiao Zhang[2], Xuewei Cao[3], Zhengming Ding[4], Qiuying Sha[3], Hui Shen[2], Hong-Wen Deng[2], Weihua Zhou[1,5*]

1. Department of Applied Computing, Michigan Technological University, 1400 Townsend Dr, Houghton, MI 49931, USA

2. Division of Biomedical Informatics and Genomics, Tulane Center of Biomedical Informatics and Genomics, Deming Department of Medicine, Tulane University, New Orleans, LA 70112, USA

3. Department of Mathematical Sciences, Michigan Technological University, 1400 Townsend Dr, Houghton, MI 49931, USA

4. Department of Computer Science, Tulane University, New Orleans, LA 70118, USA

5. Center for Biocomputing and Digital Health, Institute of Computing and Cybersystems, and Health Research Institute, Michigan Technological University, Houghton, MI 49931, USA

* Corresponding authors and lead contacts:

Weihua Zhou, Ph.D.

Department of Applied Computing,

Michigan Technological University,

1400 Townsend Dr,

Houghton, MI, 49931, USA

Tel: 906-487-2666 E-Mail: whzhou@mtu.edu

Hong-Wen Deng, Ph.D.

Tulane Center for Biomedical Informatics and Genomics,

Deming Department of Medicine,

Tulane University,

1440 Canal Street, Suite 1619F,

New Orleans, LA 70112, USA

Tel: 504-988-1310 Email: hdeng2@tulane.edu

**Abstract**

Integration of heterogeneous and high-dimensional multi-omics data is becoming increasingly important in understanding genetic data. Each omics technique only provides a limited view of the underlying biological process and integrating heterogeneous omics layers simultaneously would lead to a more comprehensive and detailed understanding of diseases and phenotypes. However, one obstacle faced when performing multi-omics data integration is the existence of unpaired multi-omics data due to instrument sensitivity and cost. Studies may fail if certain aspects of the subjects are missing or incomplete. In this paper, we propose a deep learning method for multi-omics integration with incomplete data by Cross-omics Linked unified embedding with Contrastive Learning and Self Attention (CLCLSA). Utilizing complete multi-omics data as supervision, the model employs cross-omics autoencoders to learn the feature representation across different types of biological data. The multi-omics contrastive learning, which is used to maximize the mutual information between different types of omics, is employed before latent feature concatenation. In addition, the feature-level self-attention and omics-level self-attention are employed to dynamically identify the most informative features for multi-omics data integration. Extensive experiments were conducted on four public multi-omics datasets. The experimental results indicated that the proposed CLCLSA outperformed the state-of-the-art approaches for multi-omics data classification using incomplete multi-omics data.

# 1. Introduction

The development of high-throughput omics technologies has revolutionized our ability to study biological systems at a molecular level [1]. These high-throughput techniques, including genomics, transcriptomics, proteomics and epigenomics, allow us to profile the genetic expression and interaction of molecules from different biological perspectives [2]. However, each omics technique only provides a limited view of the underlying biological process. Integrating heterogeneous omics layers simultaneously would lead to a more comprehensive and detailed understanding of diseases and phenotypes.

Numerous methods for integrating multi-omics data have been proposed, including neural network-based integration [3], machine learning-based integration [4], and pathway-based integration [5]. Neural network-based integration involves constructing networks for each omics data type and then integrating them to generate a more comprehensive network that captures the interactions between different omics layers, such as mRNA-miRNA interaction [6]. Machine learning-based integration uses machine learning models to integrate different types of omics data [2]. Pathway-based integration involves mapping omics data into known biological pathways to identify key pathways that are dysregulated in diseases [5]. However, currently available integration methods for multi-omics data mostly rely on unreliable approaches, such as simply concatenating the latent feature representation [7] or raw input data [8] from different types of omics data, which limits the potential for computer aided diagnosis using multi-omics data [9].

Due to the high cost or limited sensitivity of instruments, it is possible that one or more omics data types of a biological sample may be missing. The presence of partially missing individual-level observations poses a major challenge in multi-omics data integration [10,11]. A general solution to the multi-source data integration is to map the data from different omics layers into a common space and employing single-omics data integration method for downstream analysis [12,13]. However, explicitly projecting the features from different omics types results in information loss due to the heterogeneity of multi-omics data [14] and further decreases the performance of downstream tasks, making it challenging yet important to effectively utilize incomplete omics data. Another general solution is to use complete case which considers only the set of subjects with complete observations across all omics data [15]. This approach is convenient; however, it results in a decrease in sample size and the trained model only reflects the performance on the partial dataset. To fully use the entire dataset with incomplete multi-omics data, the multi-omics imputation based methods perform the data completion on raw input [16], which is challenging because of the high dimension of multi-omics data. Thus, an effective method, which could fully use partial multi-omics data, is necessary to integrate incomplete multi-omics data.

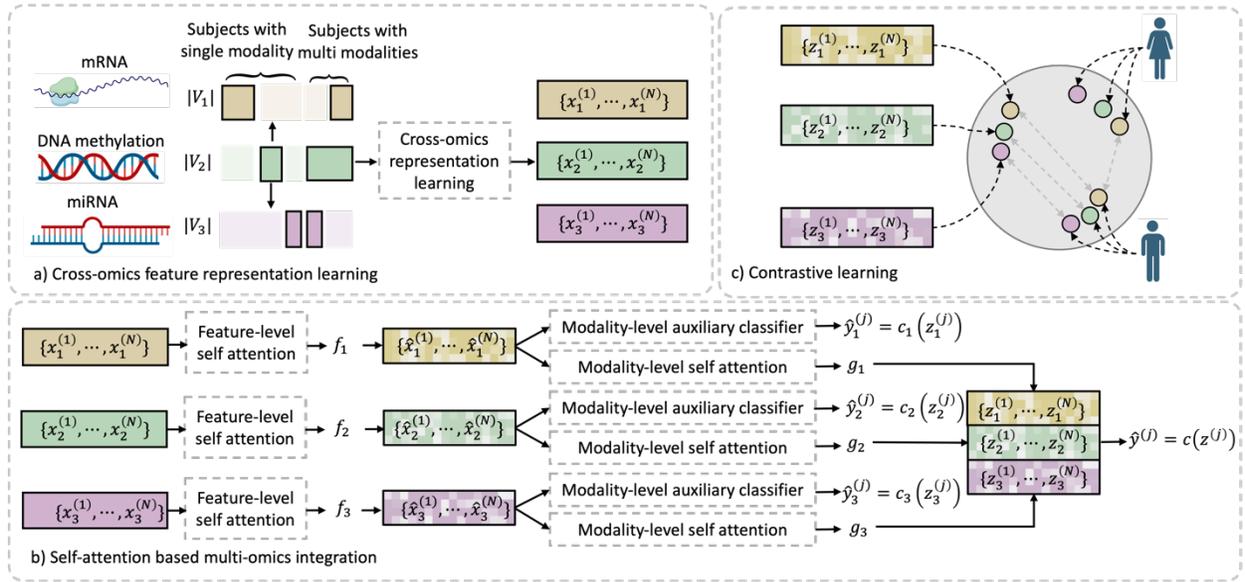

**Figure 1**. Architecture of the proposed Cross-omics Linked unified embedding with Contrastive Learning and Self Attention (CLCLSA). a) cross-omics feature representation learning. To perform multi-omics integration using incomplete multi-omics data, CLCLSA first performs the cross-omics feature representation prediction using existing omics data. Our cross-omics feature representation learning module learns the complete set of omics layers for downstream tasks. Each subject has at least one omics layer, and the solid rectangle with black borders indicate that the omics data is included while the solid rectangle without black borders means that the omics data is missing. b) self-attention-based omics-specific feature embedding and cross-omics feature integration. The embedded features from each omics are concatenated for multi-omics data classification. C) contrastive learning for consistency learning. In this example, three types of omics data, including mRNA expression, DNA methylation and miRNA expression, are included for cancer classification with incomplete omics layers.

**Contribution**: In this paper, we propose a deep learning method for multi-omics integration with incomplete data by Cross-omics Linked unified embedding with Contrastive Learning and Self Attention (CLCLSA), which contains three key components, as shown in Figure 1. The novelty and contribution of this study are as follows.

1) Utilizing complete multi-omics data as supervision, the model uses cross-omics autoencoders to learn the cross-omics feature representation. With cross-omics embedding, CLCLSA can reconstruct the incomplete multi-omics data by calculating the modality-specific representation.

2) The multi-omics contrastive learning is used to maximize the mutual information between different omics layers as shown in in Figure 1 (c).

3) Feature-level self-attention and omics-level self-attention are employed to dynamically select the most informative features for multi-omics data integration. The cross-omics autoencoders are used to restore missing omics data and the contrastive learning and self-attention are used to boost model performance.

Extensive experiments were performed on four public datasets for multi-omics data classification. Experimental results indicated that the CLCLSA achieved the state-of-the-art performance on these tasks using both complete and incomplete multi-omics data. Our code is available in https://github.com/MIILab-MTU/CLCLSA.

## 2. Related work

## 2.1. Multi-view integration

Multi-view learning (MVL) is a strategy for fusing data from different modalities. MVL approaches can be divided into three categories: 1) co-training, 2) multi-kernel learning, and 3) subspace learning [17]. Co-training exchanges the separately trained models and provides the pseudo labels with discriminative information between views to accelerate model training, which focuses more on semi-supervised learning tasks [18]. Multi-kernel learning aims at finding the mapping between different views with different kernels, and then combines the projected features for information fusion [19]. Subspace learning aims at finding the common space between different views where the shared latent space contains the information of all views [20]. For subspace learning, autoencoder is widely used for extracting the common representation of the multi-view data.

Machine learning and artificial intelligence have also shown promising results in multi-omics integration tasks. Xie et al. performed early concatenation of raw features from different omics and employed a deep neural network for implicit feature embedding [8]. Wang et al. [2] developed a deep learning based method for multi-omics data integration using graph convolutional network (GCN) for within omics feature integration and view correlation discovery network for cross-omics feature integration. Han et al. developed a dynamic fusion method for trustworthy multi-omics data classification, which dynamically calculated the feature importance and modality importance for omics data integration [9]. Although these methods achieved impressive performance for multi-omics data integration, they were designed for complete multi-omics data integration.

## 2.2. Incomplete multi-view representation learning

Missing data are common in generating multi-omics data. It can arise due to a variety of reasons, such as insufficient sample volume, instrument sensitivity, data collection cost and poor tissue quality [21]. Most of the incomplete multi-view integration methods adopt the two-stage training strategy, which constructs complete multi-view observations based on imputation methods [22,23] and performs multi-omics integration based on the existing multi-view data and the imputed data. Lee employed the variational information bottleneck approach to integrate incomplete multi-omics data for classification [24]. Multi-omics factor analysis (MOFA) [25] and MOFA+ [26] are unsupervised integrative approaches that factorize each omics matrix into the products of two components and the shared component is used to generate latent representation for downstream tasks. Multimodal variational autoencoder was proposed as an unsupervised approach for integrative multi-view analysis [27]. Though these unsupervised learning methods are capable of learning latent representation, the task-relevant information may be lost because the labels of the biological samples are not used during the training. Compared to existing incomplete multi-view integration methods, our proposed CLCLSA not only imputes the missing omics using unsupervised learning, but also uses the supervised learning to guide the imputation module to generate a precise latent representation for missing data. In addition, our CLCLSA imputes the missing omics data on latent space rather than the input space, which reduces the dimensionality of the imputation task and improves the quality of the imputed data.

## 2.3. Contrastive learning

Contrastive learning is one of the most effective unsupervised learning methods in the field of representation learning. Contrastive learning aims at seeking a latent space that separates the pairs from different classes and maximizes the similarity for samples with the same classes [28]. Existing methods maximize the mutual information between the augmented data pairs [28,29] and adopt InfoNCE as the contrastive loss. However, our method uses the multi-view omics data as the compared pairs to maximize the consistency across different omics data.

## 3. Methodology

We assume that there are $N$ subjects with $M$ omics data to be integrated. Each subject has a distinct feature set $\mathcal{X}_i = \{x_i^{(1)}, x_i^{(2)}, \cdots, x_i^{(N)}\}$, where $x_i^{(j)} \in \mathbb{R}^{|V_i|}$ represents the omics features for $j$-th subject of $i$-th omics; $|V_i|$ is the feature dimension of $i$-th omics. To avoid the ambiguity between the index of the omics layers and subjects, we use $i \in \{1, \cdots, M\}$ in the subscript to represent the index of omics layer and $j \in \{1, \cdots, N\}$ in the superscript to represent the index of the subject. The overview of our proposed CLCLSA is shown in Figure 1.

### 3.1. Self-attention based dynamical integration

**Feature-level self-attention.** Inspired by Han et al.[9], we employ the self-attention component for feature-level feature selection and omics-level feature selection. Though sparsity of the high-dimensional data enables a high-level feature embedding, the informativeness of different features among different subjects are more important for multi-omics integration [30]. The calculated feature-level attention scores are denoted as

$$fatt_i^{(j)} = \sigma\left(f_i\left(x_i^{(j)}\right)\right) \tag{1}$$

where $f_i$ is the feature-level self-attention encoder implemented by multi-layer perceptron (MLP) and $\sigma$ is the sigmoid activation function. The features are further embedded by omics-specific encoder $emb_i$ and the calculated feature-level attention scores, which is denoted as

$$\hat{x}_i^{(j)} = emb_i\left(x_i^{(j)} \odot fatt_i^{(j)}\right) \tag{2}$$

where $\odot$ represents the element-wise multiplication.

**Omics-level self-attention.** The importance of different modalities is not fixed during the multi-omics fusion [31]. We further employ the omics-level self-attention encoder $g_i$ to calculate the omics-level importance, which is denoted as

$$matt_i^{(j)} = \sigma\left(g_i\left(\hat{x}_i^{(j)}\right)\right) \tag{3}$$

where $g_i$ is implemented using MLP. The latent features $\hat{x}_i^{(j)}$ is further embedded by omics-level attention scores, as shown in Eq. 4.

$$\hat{z}_i^{(j)} = \hat{x}_i^{(j)} \odot matt_i^{(j)} \tag{4}$$

**Multi-view fusion.** A nested multi-omics fusion is employed in this study which concatenates the features performed by feature-level self-attention and omics-level self-attention, as shown in Eq. 5.

$$z^{(j)} = [\hat{z}_1^{(j)}, \cdots, \hat{z}_M^{(j)}] \tag{5}$$

where $[\cdot]$ indicates the concatenation operator and $z$ is the multi-omics representation.

### 3.2. Missing omics completion

The proposed CLCLSA is a deep generative model that aligns feature representation from different omics after building a partially paired multi-omics data. The key component of CLCLSA for missing omics completion is that it models the shared latent feature representation as a combination of modality specific sub-spaces of all existing modalities. Instead of completing missing modality in the raw feature space, the cross-omics representation learning is employed to complete the missing omics in the latent feature space.

Suppose the latent feature representation for $i$-th omics is $\hat{z}_i^{(j)} \in \mathbb{R}^{D_i}$, where $D_i$ is the dimension of the latent space. In practice, we set $D_i$, s.t. $i \in \{1, \cdots, M\}$ has the same dimensions. Suppose the indices of missing omics and the known omics for $j$-th subject are $i$ and $k$, we employ the autoencoder to perform cross-omics representation learning, which is defined in Eq. 6.

$$h_{ik}\left(\hat{z}_k^{(j)}\right) = dec_i\left(enc_k\left(\hat{z}_k^{(j)}\right)\right) \tag{6}$$

where $enc_k(\cdot)$ indicates the encoder for the $k$-th omics layer and $dec_i(\cdot)$ indicates the decoder for the $i$-th omics layer; $h_{ik}\left(\hat{z}_k^{(j)}\right)$ is the predicted latent embedding of the $j$-th subject for the $i$-th omics data using the $k$-th omics data. The mean squared errors between the predicted latent representation from other omics, i.e. $h_{ik}\left(\hat{z}_k^{(j)}\right)$, and the extract feature embedding from the $i$-th omics, i.e. $\hat{z}_i^{(j)}$, is used to optimize the weights for the autoencoder $h_{ik}$, as shown in Eq. 7.

$$L_{co}^{ik}(j) = \left\|h_{ik}\left(\hat{z}_k^{(j)}\right) - \hat{z}_i^{(j)}\right\|_2^2 \tag{7}$$

Under the bi-view setting, the cross-omics data reconstruction loss is defined as

$$L_{co} = \sum_{j=1}^{N}\left(L_{co}^{ik}(j) + L_{co}^{ki}(j)\right) = \sum_{j=1}^{N}\left\|h_{12}\left(\hat{z}_2^{(j)}\right) - \hat{z}_1^{(j)}\right\|_2^2 + \left\|h_{21}\left(\hat{z}_1^{(j)}\right) - \hat{z}_2^{(j)}\right\|_2^2 \tag{8}$$

Under the multi-view setting, the cross-omics data reconstruction loss is defined in a fully connected fashion in Eq. 9.

$$L_{co} = \sum_{j}^{N}\sum_{i}^{M}\sum_{k}^{M}\mathbb{I}_{\{i \neq k\}}L_{co}^{ik}(j) = \sum_{j=1}^{N}\sum_{i=1}^{M}\sum_{k=1}^{M}\mathbb{I}_{\{i \neq k\}}\left\|h_{ik}\left(\hat{z}_k^{(j)}\right) - \hat{z}_i^{(j)}\right\|_2^2 \tag{9}$$

where $\mathbb{I}_{\{i \neq k\}}$ is an indicator function. However, optimizing cross-omics loss of Eq. 8 or Eq. 9 leads to trivial solutions that $\{\hat{z}_1^{(j)}, \hat{z}_2^{(j)}, \cdots, \hat{z}_M^{(j)}\}$ converge to the same solution [32,33]. To overcome this issue, we train the CLCLSA using both subjects with complete omics data and subjects with missing omics data. For the subjects with complete omics data, we feed them into the network to generate the feature representation for all views and train the autoencoders $h_{ik}, i \neq k$ and $i, k \in \{1 \cdots M\}$ between each of the two omics layers. For the subjects with incomplete omics, we feed them into the network to generate the feature representation for the views with omics data and use the trained $h_{ik}$ to generate the feature representation to perform the missing omics completion. Using both the complete representation, i.e., $\hat{z}_i^{(j)}$, and the predicted representation, i.e., $h_{ik}\left(\hat{z}_k^{(j)}\right)$, we fuse them as the multi-omics data representation for downstream tasks. The detailed architecture of the autoencoder is illustrated in section 4.1.

### 3.3. Contrastive learning

Contrastive learning is a widely used technique in deep learning for incomplete multi-modality fusion, which guarantees the consistency between different modalities [32]. Most of the contrastive learning algorithms maximize the lower bound of the mutual information between the augmented samples and the raw training samples [29,34] using single-view data. However, we maximize the lower bound using the mutual information between different omics layers. We adopt the contrastive learning loss function proposed in [32] to maximize the entropy of the feature representation for each omics layer as well as the mutual information between different omics layers. The cross-view contrastive learning loss is defined as:

$$L_{cl}(Z_i, Z_k) = -\sum_{d=1}^{D_i}\sum_{d'=1}^{D_k} P_{dd'} \ln \frac{P_{dd'}}{P_d^{\alpha+1} \cdot P_{d'}^{\alpha+1}} \tag{10}$$

where $Z_i = \{z_i^{(1)}, z_i^{(2)}, \cdots, z_i^N\} \in \mathbb{R}^{N \times D_k}$ is the latent representation matrix for $N$ subjects on $i$-th omics, $P \in \mathbb{R}^{D_i \times D_k}$ represents the joint probability distribution matrix of $Z_i$ and $Z_k$, as defined in Eq. 11.

$$P = \frac{1}{N}\sum_{j=1}^{N} \hat{z}_i^{(j)} \cdot \left(\hat{z}_k^{(j)}\right)^T \tag{11}$$

where $P_{dd'}$ is the $(d, d')$-th element of $P$; $P_d$ and $P_{d'}$ indicate the marginal probability distribution; and $\alpha$ is the hyperparameter balancing the cross-omics mutual information and omics-specific entropy.

For the multi-omics contrastive learning, we average the contrastive loss between each of two omics layers, as shown in Eq. 12.

$$L_{cl} = \sum_{i=1}^{M}\sum_{k=1}^{M} \mathbb{I}_{\{i \neq k\}} L_{cl}(Z_i, Z_k) \tag{12}$$

where $\mathbb{I}_{\{i \neq k\}}$ is an indicator function.

### 3.4. Loss function and training strategy

Given an incomplete multi-omics dataset $\{\mathcal{X}_i\}_{i=1}^{M}$ and the classification gold standards $\{y^{(j)}\}_{j=1}^{N}$, we first feed the subjects with complete multi-omics data into the CLCLSA. For the subjects with missing omics, CLCLSA completes the latent feature representation of the missing omics layer(s) by cross-omics autoencoders defined in Section 3.2. The concatenated features are used for classification and the classifier is denoted as $c$, as shown in Figure 1 (b). In addition, for each omics layer, an auxiliary classifier is employed to boost model training, which is denoted as $c_i$ for the $i$-th omics layer, as shown in Figure 1 (b).

During the model training, we set the cross-entropy loss as the loss for the classifier $c$, as shown in Eq. 13.

$$L_{clf} = -\sum_{j=1}^{N} y^{(j)} \log \hat{y}^{(j)} \tag{13}$$

where $\hat{y}^{(j)} = c(z^{(j)})$ is the classifier output.

Since both of the omics attention score and omics auxiliary classifier probability are the maximal Softmax outputs, they promisingly reflect the classification confidence [9]. We adopt the auxiliary classification loss for each omics as a regularity for both omics-specific attention encoder $g_i$ and omics-specific auxiliary classifier $c_i$, as defined in Eq. 14.

$$L_{al} = \sum_{i}^{M} \left(matt_i^{(j)} - \hat{y}_i^{(j)}\right)^2 + y^{(j)} \log \hat{y}_i^{(j)} \tag{14}$$

where $\hat{y}^{(j)} = c_i\left(z_i^{(j)}\right)$ is the auxiliary classifier output.

The overall loss function for CLCLSA is defined in Eq. 15.

$$L = L_{clf} + \lambda_{al}L_{al} + \lambda_{co}L_{co} + \lambda_{cl}L_{cl} \tag{15}$$

where $\lambda_{al}$, $\lambda_{co}$ and $\lambda_{cl}$ are three hyperparameters balancing the weights of the auxiliary classification loss, cross-omics data reconstruction loss and multi-omics contrastive loss.

## 4. Experiments and discussion

To validate the proposed CLCLSA model, we applied the proposed model on four empirical multi-omics datasets. Extensive experimental results indicated the superiority of the proposed method on multi-omics data classification tasks using both complete and incomplete multi-omics data. In addition, we also analyzed the hyperparameter settings and performed ablation studies to demonstrate the effectiveness of different components.

### 4.1. Datasets and comparison approaches

We conduct experiments on the following four widely used multi-omics datasets. For each subject in these four datasets, the mRNA expression, DNA methylation and miRNA expression data were used. We adopt the same data preprocessing pipeline as [2], and set the same feature dimension for each omics. To fairly compare the proposed CLCLSA with existing approaches, we follow the same protocol of [2,9] to build the training set and testing set for evaluation.

- **ROSMAP** dataset [35,36], which contains 351 subjects, is used for distinguishing Alzheimer's Disease (AD) subjects from normal controls.
- **LGG** dataset [37] is for low-grade glioma classification, which contains 246 grade-2 subjects and 264 grade-3 subjects.
- **BRCA** dataset [38] is for breast invasive carcinoma PAM50 subtype classification, which contains 876 subjects among 5 classes.
- **KIPAN** dataset [38] is used for kidney cancer type classification, which contains 658 subjects from 3 classes.

For the performance evaluations using complete multi-omics datasets, we compared the proposed method with the following 13 existing classification algorithms:

- Five machine learning algorithms. K-nearest neighbor classifier (**KNN**) [39], Support Vector Machine (**SVM**) [40], Linear Regression with L1 regularization (**LR**), Random Forest (**RF**) [41] and fully connected neural networks (**NN**).
- Adaptive group-regularized ridge regression (**GRridge**) [42]. GRridge is a method for adaptive group-regularized ridge regression with group-specific penalties for high-dimensional data classification.
- Block partial least squares discriminant analysis (**BPLSDA**) and Block sparse partial least squares discriminant analysis (**BSPLSDA**) [43]. BPLSDA extends sparse generalized canonical correlation analysis to a classification framework, and BSPLSDA adds sparse constraints to BPLSDA.
- Multi-Omics Graph cOnvolutional NETworks (**MOGONET**) [2]: MOGONET jointly explores omics-specific learning using graph convolution network and cross-omics correlation learning using view correlation discovery network for multi-omics integration and classification.
- Trusted multi-view classification (**TMC**) [44]. TMC dynamically computes the trustworthiness of each modality for different subjects with reliable integration for multi-view classification.
- Concatenation of final multimodal representations (**CF**) [45]. CF performs multi-view information fusion based on late fusion and compactness-based fusion.

- Gated multimodal units for information fusion (**GMU**) [46]. GMU employs the gates for selecting the most important parts of the input of each modality to correctly generate the desired output.
- Multi-modality dynamic fusion (**MMDynamic**) [9]. MMDynamic employs the feature-level and modality-specific gates for multi-modality data fusion.

For the performance evaluation using incomplete multi-omics datasets, we compared the proposed method with the following 7 existing multi-modality fusion algorithms:

- Multiview canonical correlation analysis (**MCCA**) [47]. MCCA extends the canonical correlation analysis (CCA) into multi-view settings. CCA is a typical subspace learning algorithm, aiming at finding the pairs of projections from different views with the maximum correlations. For more than 2 views, MCCA optimizes the sum of pairwise correlations. For samples with missing modalities, the trained MCCA projects the existing modalities into the subspace as the feature representations.
- Kernel CCA (**KCCA**) [48]. KCCA extends MCCA by adding kernel techniques.
- Kernel generalized CCA (**KGCCA**) [49]. KGCCA extends KCCA with a prior-defined graph between different modalities.
- Sparse CCA (**SCCA**) [50]. SCCA extends CCA with modality-specific sparse penalty.
- Multi-view Variation AutoEncoder (**MVAE**) [27,51]. MVAE extends variational autoencoders for latent feature extraction and employs the product of the multivariate Gaussian distributions for information fusion. For samples with missing modalities, the trained MVAE generates latent representation by the product of the latent features from the existing modalities.
- Cross Partial Multi-View Networks (**CPM**) [52]. CPM performs feature embedding using incomplete multi-view data by focusing on the completeness and versatility of the feature embedding.
- Dual Contrastive Prediction (**DCP**) [32]. DCP employs maximizes the conditional entropy through dual prediction to recover the missing views and employs dual contrastive loss to learn the consistent representation among different modalities.

For the above baseline models using the complete multi-omics datasets, we compared the CLCLSA with the performance reported in [2,9]. For the models using the incomplete multi-omics datasets, we also performed the grid search to find the optimal hyperparameters and network settings and compared the performance from the best models. To evaluate the CLCLSA performance in handling incomplete data, we manually created data with different missing rates. The missing rate is denoted as $\eta = N_{ic}/N$, where $N_{ic}$ represents the number of subjects with at most $M - 1$ omics layers.

**Evaluations**. To evaluate the performance between the proposed CLCLSA and comparison approaches for binary classification on ROSMAP and LGG datasets, we employed accuracy (ACC), F1-score (F1) and area under the receiver operating characteristic curve (AUC) to evaluate the performance. To evaluate the multi-class classification performance, we employed the ACC, weighted F1 score (WeightedF1) and macro-averaged F1 score (MacroF1).

### 4.2. Network architecture and implementation details

The proposed CLCLSA contains 6 components, i.e., the feature-level self-attention encoder $f_i$, the omics-specific encoder $emb_i$, the omics-level self-attention encoder $g_i$, the omics-specific auxiliary classifier $c_i$, the final classifier $c$, and the omics-to-omics conditional sparse autoencoder $h_{ik}$. The $f_i$ and $g_i$ were implemented using MLPs. The $emb_i$ was implemented by MLPs followed by ReLU activation functions and Dropout layer (DP) with the dropout probability of 0.5. The $h_{ik}$ was implemented by MLPs followed by batch normalization (BN) layer [53] and ReLU activation functions. The $c_i$ and $c$ were implemented by MLPs and Softmax classifiers. The feature dimensions for these four multi-omics datasets, and the settings of $f_i$, $emb_i$, $g_i$, $h_{ik}$, $c_i$ and $c$ are shown in Table 1.

**Table 1.** Feature dimensions of the four multi-omics datasets and the network settings of our CLCLSA. The feature dimensions represent the dimension of mRNA expression, DNA methylation and miRNA expression sequentially. The network settings are for the mRNA expression, DNA methylation and miRNA omics layers, respectively.

| Dataset | ROSMAP | LGG | BRCA | KIPAN |
|---|---|---|---|---|
| Categories | 2 | 2 | 5 | 5 |
| Feature dimensions | 200, 200, 200 | 2000, 2000, 548 | 1000, 1000, 503 | 2000, 2000, 445 |
| $f_i$ | 200-200, 200-200, 200-200 | 2000-2000, 2000-2000, 548-548 | 1000-1000, 1000-1000, 503-503 | 2000-2000, 2000-2000, 445-445 |
| $emb_i$ | 200-300-ReLU-DP, 200-300-ReLU-DP, 200-300-ReLU-DP | 2000-200-ReLU-DP, 2000-200-ReLU-DP, 548-200-ReLU-DP | 1000-200-ReLU-DP, 1000-200-ReLU-DP, 503-200-ReLU-DP | 2000-200-ReLU-DP, 2000-200-ReLU-DP, 445-200-ReLU-DP |
| $g_i$ | 300-1, 300-1, 300-1 | 200-1, 200-1, 200-1 | 200-1, 200-1, 200-1 | 200-1, 200-1, 200-1 |
| $h_{ik}$ | 300-64-BN-ReLU-32-ReLU-64-BN-ReLU-300 | 200-64-BN-ReLU-32-ReLU-64-BN-ReLU-200 | 200-64-BN-ReLU-32-ReLU-64-BN-ReLU-200 | 200-64-BN-ReLU-32-ReLU-64-BN-ReLU-200 |
| $c_i$ | 300-2-Softmax, 300-2-Softmax, 300-2-Softmax | 200-2-Softmax, 200-2-Softmax, 200-2-Softmax | 200-5-Softmax, 200-5-Softmax, 200-5-Softmax | 200-3-Softmax, 200-3-Softmax, 200-3-Softmax |
| $c$ | 900-2-Softmax | 600-2-Softmax | 600-5-Softmax | 600-5-Softmax |

**Grid search**. To investigate the effectiveness of different components of the proposed CLCLSA, we performed the hyperparameter finetuning by grid search. In detail, we trained the CLCLSA on four datasets using different $\lambda_{al}$, $\lambda_{co}$ and $\lambda_{cl}$, where each hyperparameter was set as one of {0, 0.01, 0.02, 0.05, 0.1, 1.0}, '0' indicated the corresponding component was removed during the model training. Note that the $\lambda_{co}$ was set as 0 when training with the complete multi-omics data and $\lambda_{co}$ was set greater than 0 when training with the incomplete multi-omics data. For the models trained on different datasets and different missing rates, the optimal hyperparameters varied.

We implemented the CLCLSA using PyTorch 1.10 [54], and the experiments were carried on a workstation with an NVIDIA RTX A6000 GPUs. Adam optimizer [55] with the initial learning rate of 0.0001 and the learning rate decay technique was employed to optimize the model weights. The maximum training epochs were set fixed as 2,500 for all experiments. The batch size was set as 245, 357, 612 and 460 for ROSMAP, LGG, BRCA and KIPAN datasets. The training times were 80s, 75s, 100s and 85s for ROSMAP, LGG, BRCA and KIPAN datasets, respectively.

### 4.3. Performance using complete multi-omics data

We compared our CLCLSA model with existing multi-view classification algorithms, as shown in Table 2. When using the complete multi-omics data, the cross-omics autoencoders were removed and the $L_{co}$ was set as zero.

**Table 2.** Comparison with the state-of-the-art algorithms for multi-omics data classification. The bold texts indicate the best performance.

| Dataset | ROSMAP | | | LGG | | | BRCA | | | KIPAN | | |
|---|---|---|---|---|---|---|---|---|---|---|---|---|
| method | ACC | F1 | AUC | ACC | F1 | AUC | ACC | WeightedF1 | MacroF1 | ACC | WeightedF1 | MacroF1 |
| KNN | 65.7 | 67.1 | 70.9 | 72.9 | 73.8 | 79.9 | 74.2 | 73.0 | 68.2 | 96.7 | 96.7 | 96 |
| SVM | 77.0 | 77.8 | 77.0 | 75.4 | 75.7 | 75.4 | 72.9 | 70.2 | 64.0 | 99.5 | 99.5 | 99.4 |
| LR | 69.4 | 73.0 | 77.0 | 76.1 | 76.7 | 82.3 | 73.2 | 69.8 | 64.2 | 97.4 | 97.4 | 97.2 |
| RF | 72.6 | 73.4 | 81.1 | 74.8 | 74.2 | 82.3 | 75.4 | 73.3 | 64.9 | 98.1 | 98.1 | 97.5 |
| NN | 75.5 | 76.4 | 82.7 | 73.7 | 74.8 | 81.0 | 75.4 | 74.0 | 66.8 | 99.1 | 99.1 | 99.1 |
| GRridge | 76.0 | 76.9 | 84.1 | 74.6 | 75.6 | 82.6 | 74.5 | 72.6 | 65.6 | 99.4 | 99.4 | 99.3 |
| BPLSDA | 74.2 | 75.5 | 83.0 | 75.9 | 73.8 | 82.5 | 64.2 | 53.4 | 36.9 | 93.3 | 93.3 | 91.9 |
| BSPLSDA | 75.3 | 76.4 | 83.8 | 68.5 | 66.2 | 73.0 | 63.9 | 52.2 | 35.1 | 91.9 | 91.8 | 89.5 |
| MOGONET | 81.5 | 82.1 | 87.4 | 81.6 | 81.4 | 84.0 | 82.9 | 82.5 | 77.4 | **99.9** | **99.9** | **99.9** |
| TMC | 82.5 | 82.3 | 88.5 | 81.9 | 81.5 | 87.1 | 84.2 | 84.4 | 80.6 | 99.7 | 99.7 | 99.4 |
| CF | 78.4 | 78.9 | 88.0 | 81.1 | 82.2 | 88.1 | 81.5 | 81.5 | 77.1 | 99.2 | 99.2 | 98.8 |
| GMU | 77.6 | 78.4 | 86.9 | 80.3 | 80.8 | 88.6 | 80.0 | 79.8 | 74.6 | 97.7 | 97.6 | 95.8 |
| MMDynamics | 84.2 | 84.6 | **91.2** | 83.3 | 83.7 | 88.5 | 87.7 | 88.0 | 84.5 | **99.9** | **99.9** | **99.9** |
| MCCA | 58.1 | 67.2 | 57.4 | 76.3 | 72.3 | 76.7 | 72.9 | 68.7 | 56.9 | 92.4 | 91.8 | 85.2 |
| KCCA | 59.0 | 67.7 | 58.3 | 73.7 | 69.7 | 74.1 | 67.9 | 61.1 | 49.0 | 92.9 | 92.5 | 86.8 |
| KGCCA | 59.0 | 67.6 | 58.3 | 74.3 | 69.8 | 74.8 | 67.9 | 61.1 | 49.0 | 92.9 | 92.5 | 86.8 |
| SCCA | 75.2 | 76.3 | 75.2 | 77.6 | 77.1 | 77.8 | 76.0 | 74.3 | 68.2 | 91.9 | 91.5 | 86.0 |
| MVAE | 79.0 | 78.4 | 79.2 | 77.6 | 77.7 | 77.8 | 84.7 | 84.7 | 81.5 | 93.4 | 93.3 | 92.2 |
| CPM | 73.3 | 70.2 | 73.1 | 71.7 | 76.8 | 72.3 | 82.4 | 82.8 | 79.6 | 98.0 | 98.0 | 98.5 |
| DCP | 69.1 | 66.2 | 68.9 | 72.8 | 73.1 | 72.3 | 79.1 | 79.7 | 73.5 | 97.0 | 97.0 | 95.4 |
| CLCLSA | **85.7** | **85.3** | 81.7 | **86.3** | **86.1** | **90.5** | **88.6** | **89.0** | **86.6** | **99.9** | **99.9** | **99.9** |

For the binary classification tasks, we compared the CLCLSA with the existing methods on ROSMAP and LGG datasets. Experimental results demonstrated that the proposed CLCLSA outperformed other methods on binary classification significantly. For the ROSMAP dataset, the CLCLSA improved the ACC from 84.2%, achieved by the second-best algorithm, to 85.7%; for LGG dataset, the CLCLSA improved ACC from 83.3% achieved by the second-best algorithm to 86.3%. The results indicated that the contrastive learning improved the differentiability of feature representation among different classes so that promoted the model performance.

For the multi-classification tasks, we compared the CLCLSA with the existing methods on BRCA and KIPAN datasets. For the BRCA dataset, the proposed method slightly improved the ACC from 87.7% to 88.6%. The difference between our method and MMDynamics is that our CLCLSA employs the contrastive learning, which increases the mutual information between different omics and promotes the differentiability between subjects, resulting a higher performance. For the KIPAN dataset, the proposed method achieved the same performance with the MMDynamics [9] and MOGONET [2]. Classifying kidney cancer type using KIPAN dataset was a relatively simpler task that all methods achieved quite high performance.

### 4.4. Performance using incomplete multi-omics data

We further compared the proposed CLCLSA with other existing multi-view incomplete data classification methods on these four public multi-omics datasets. We manually specified the missing rates ranging from 0.1 to 0.8 with an increment of 0.1, and the performance of multi-omics data classification using incomplete multi-omics data is shown in Figure 2.

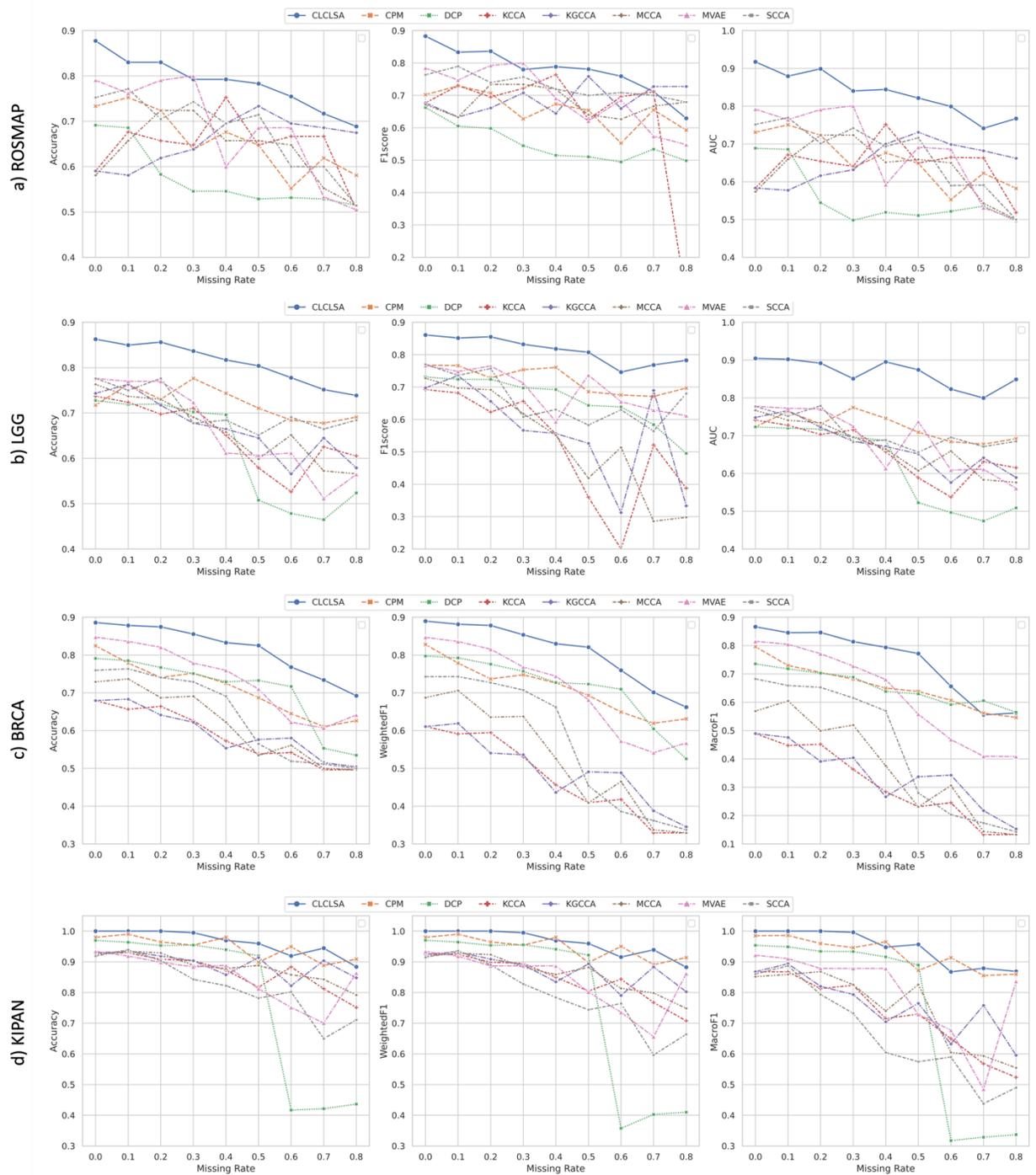

**Figure 2**. Classification performance with different missing rates on a) ROSMAP, b) LGG, c) BRCA and d) KIPAN datasets.

Figure 2 shows that the CLCLSA method outperformed all other state-of-the-art multi-view classification methods when using incomplete multi-omics data in most settings. For the binary classification task, the CLCLSA achieved the highest ACCs among most of the datasets with different missing rates.

For ROSMAP dataset, the CLCLSA achieved the AUCs of 0.8 if the missing rate was smaller than 0.6. For the performance using the multi-omics data with the missing rate of 0.3, the MVAE achieved a higher ACC and F1Score than our CLCLSA.

For LGG dataset, the CLCLSA achieved the highest performance using incomplete multi-omics data with different missing rates. Compared to CCA based methods, i.e., MCCA, KCCA, KGCCA and SCCA, the ACC, F1-score and AUC obtained by CLCLSA dropped slightly as the increase of the missing rates. CCA based methods cannot impute missing views and the results indicate that CLCLSA benefits from the missing omics completion using cross-omics autoencoders.

For multi-class classification task, the proposed CLCLSA achieved the highest ACCs and WeightedF1s among all compared methods with different missing rates on the BRCA dataset. The CCA based methods cannot recover the feature representations of missing omics layers. Since the CCA based methods were trained using partial training subjects with complete omics data, the performance dropped significantly when using incomplete multi-omics data with large missing rates. With limited training subjects, CCA methods cannot achieve a high performance with large missing rates, such as the missing rates greater than 0.5. For MVAE method, the multi-omics fusion is achieved by the product of latent distributions from the existing omics layers. Though MVAE employed 'view dropout' during the model training [56], the features for missing omics layers were not recovered which lowered the performance for the downstream classification tasks. For KIPAN dataset, the achieved ACCs of CLCLSA were higher than 88% even using the multi-omics data with the missing rate of 0.8. When the missing rate was smaller than 0.2, the performance did not decrease. The results indicated that the cross-omics autoencoders showed the capability for recovering the missing omics data using latent feature representations from the existing omics data.

We further performed the ablation study on these 4 multi-omics datasets using partial omics layers. Under the two-view setting, the cross-omics loss degraded from Eq. 9 to Eq. 8. In Figure 3, we compared the classification performance of CLCLSA with 4 types of combinations (mRNA+meth+miRNA for combining mRNA expression, DNA methylation with miRNA expression data, mRNA+miRNA for combining mRNA expression with miRNA expression data, mRNA+methy for combining mRNA expression with DNA methylation, and miRNA+methy for combining miRNA expression with DNA methylation data).

From Figure 3, we can draw the following conclusions:

1) Integrating three types of omics proved to be advantageous for classification, particularly when dealing with complete data or when the missing rate is small. For example, in the case of multi-omics data classification for LGG and BRCA datasets, integrating three types of omics was found to be beneficial when the missing rates were smaller than 0.1. Similarly, when the missing rates were smaller than 0.3, integrating three-omics layers were beneficial for multi-omics data classification on KIPAN dataset.
2) Integrating mRNA expression data can be advantageous even for multi-omics data classification when the missing rate was high. For the ROSMAP dataset, the CLCLSA models trained with mRNA expression and DNA methylation (in green dashed line) and CLCLSA models trained with mRNA expression and miRNA expression (in orange dashed line) outperformed the CLCLSA models trained using three omics layers when the missing rates were set as 0.1, 0.2, 0.3 and 0.8. However, the CLCLSA models trained with miRNA expression and DNA methylation (in red dashed line) showed inferior performance compared to the combinations of other three omics if the missing rates were set as 0 to 0.4 with an increment of 0.1. For BRCA dataset, the CLCLSA models achieved the lowest ACCs, WeightedF1 and MacroF1s using miRNA expression and DNA methylation data. These results indicated that mRNA expression data was more important than DNA methylation and miRNA expression data for multi-omics data classification on these four datasets.

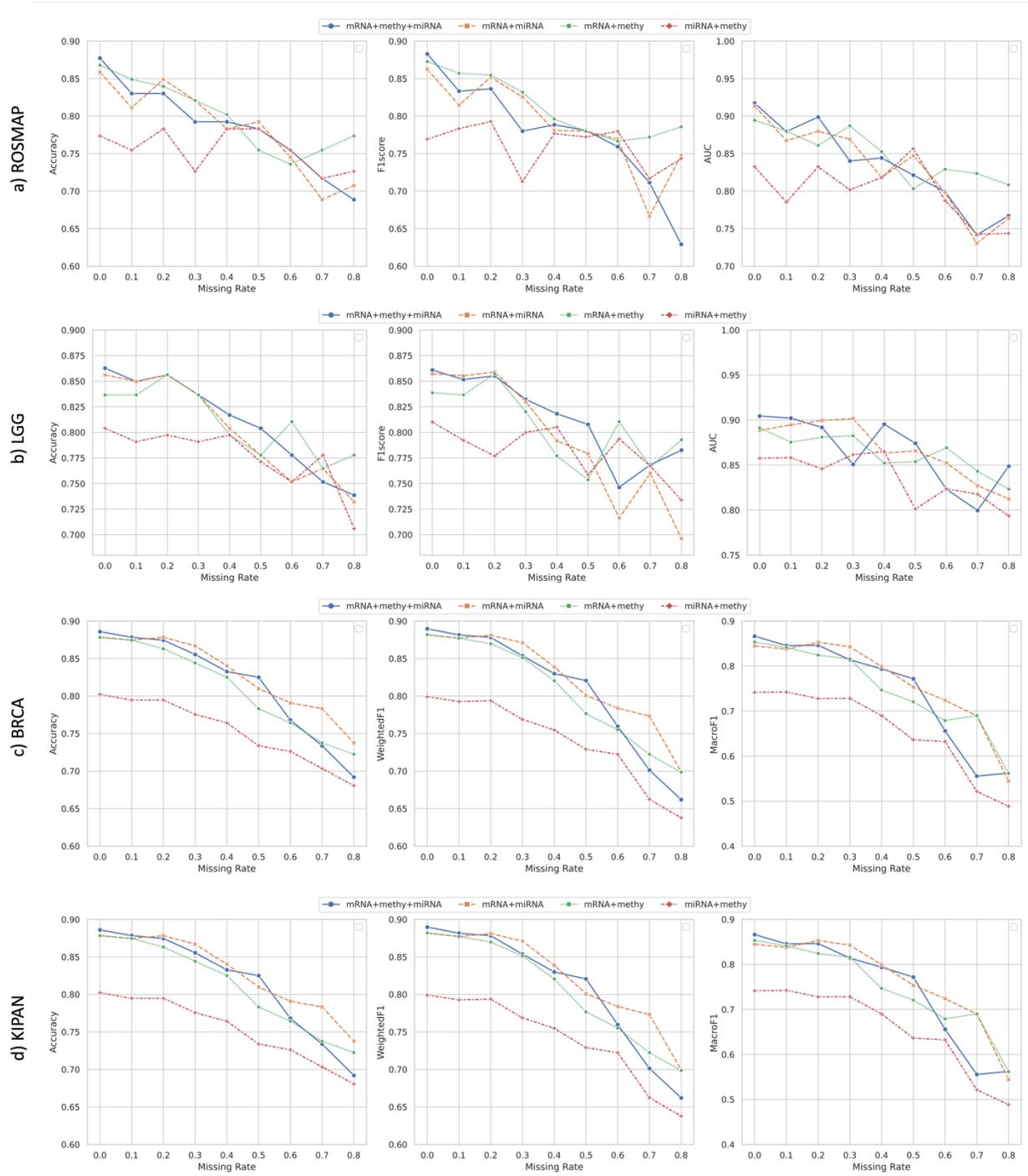

**Figure 3**. Performance of full-omics data classification and partial-omics data classification via CLCLSA. Results for a) ROSMAP, b) LGG, c) BRCA and d) KIPAN are depicted from top to bottom. mRNA+methy+miRNA refers to the classification using full-omics data. mRNA+miRNA, mRNA+methy and miRNA+methy indicates the classification using partial-omics data.

### 4.5. Hyperparameter settings and model performance

We further investigated the model performance under different hyperparameter settings, i.e. $\lambda_{al}$, $\lambda_{co}$ and $\lambda_{cl}$. We performed the comparison studies on the ROSMAP dataset with the fixed missing rate of 0.2. In

Figure 4(a), we fixed the weight of the auxiliary classification loss $\lambda_{al}$ as 0.1 and tested the model performance under different weights of contrastive loss and cross-omics completion loss; in Figure 4(b), we fixed the weight of the contrastive loss $\lambda_{cl}$ as 0.1, and tested the model performance under different weights of auxiliary classification loss and cross-omics completion loss; in Figure 4(c), we fixed the weight of the cross-omics completion loss $\lambda_{co}$ as 1, and tested the model performance under different weights of auxiliary classification loss and contrastive loss.

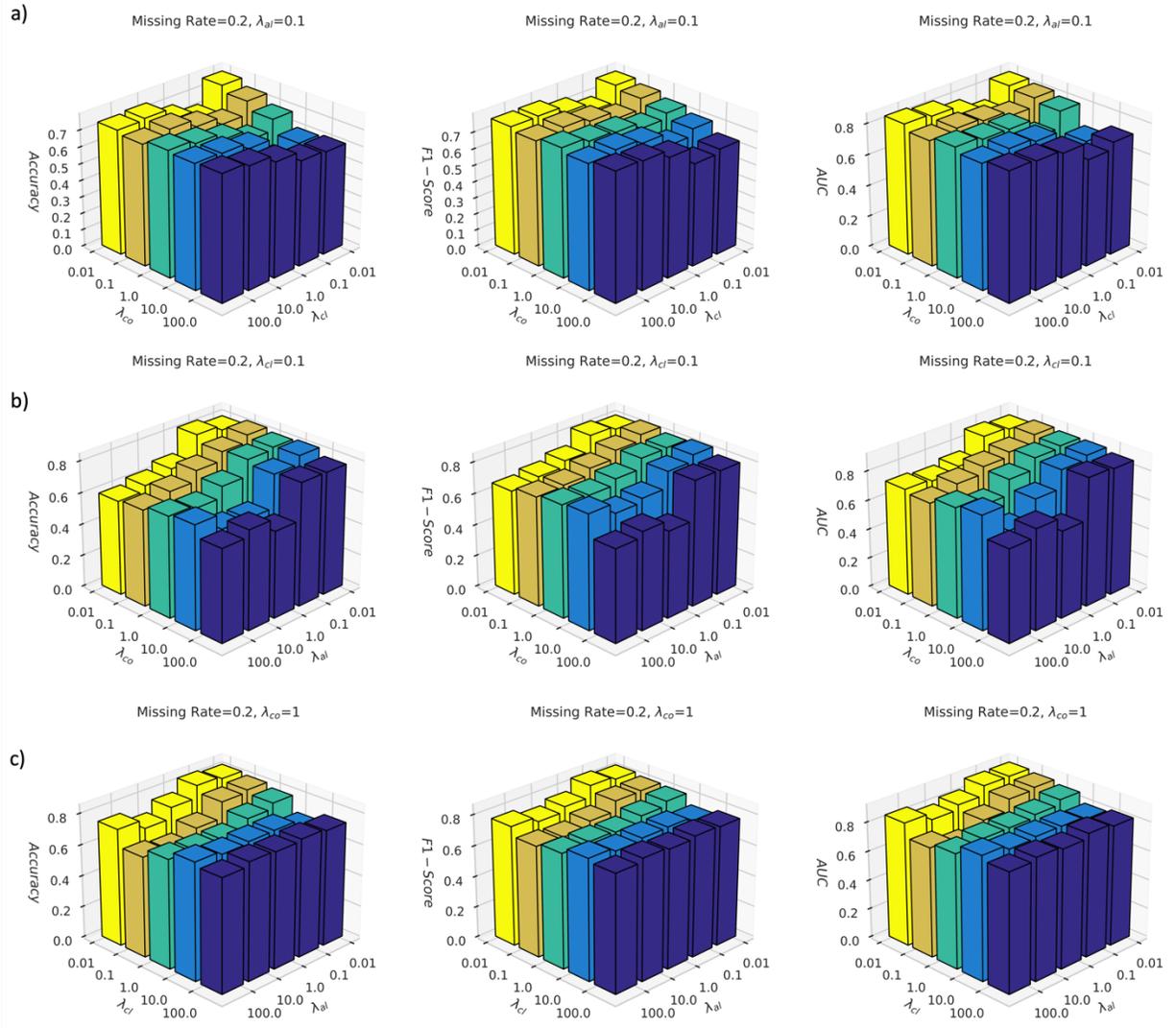

**Figure 4**. Hyperparameter analysis of CLCLSA training on the ROSMAP dataset with the fixed missing rate of 0.2. a) CLCLSA performance with the fixed balancing factor of auxiliary classification loss, $\lambda_{al} = 0.1$; b) CLCLSA performance with the fixed balancing factor of contrastive loss, $\lambda_{cl}=0.1$; c) CLCLSA performance with the fixed of balancing factor cross-omics completion loss, $\lambda_{co} = 1$.

From Figure 4 (a), it was observed that the proposed CLCLSA achieved the highest ACC, F1-Score and AUC with smaller balancing factors of $\lambda_{co}$ and $\lambda_{cl}$. When the contrastive loss balancing factor $\lambda_{cl}$ was fixed at 0.01, monotonously decreasing the balancing factor $\lambda_{co}$ increased the model performance. This indicated that setting a high balancing factor for cross-omics autoencoders decreased the performance because the cross-omics latent feature completion was not as important as contrastive learning, which increased the mutual information between different omics and promoted the differentiability between subjects. A smaller balancing factor of cross-omics prediction loss is recommended in our CLCLSA model.

From Figure 4 (b), it was observed that the proposed CLCLSA achieved the highest performance with a smaller balancing factor of $\lambda_{al}$. The 3D bar charts showed a trend that with a fixed $\lambda_{co}$, the model performance was increased in most settings with the decrease of the $\lambda_{al}$. The $\lambda_{al}$ controls the importance of the auxiliary classifiers and omics-specific self-attention encoder; however, it should not suppress the importance of the main classifier, i.e., $c$, for multi-omics data classification. A smaller balancing factor of $\lambda_{al}$ is recommend, such as $\lambda_{al} \leq 0.1$.

From Figure 4 (c), it was observed that setting a small contrastive loss factor $\lambda_{cl}$ and a small auxiliary classifier factor $\lambda_{al}$ simultaneously was beneficial for improving model performance. The performance of CLCLSA was degraded significantly when $\lambda_{cl}$ and $\lambda_{al}$ were set to 0.1 and 10.0 (in orange bars). However, when both of $\lambda_{cl}$ and $\lambda_{al}$ were set to small values simultaneously, such as 0.01 and 0.1, the performance of the CLCLSA was satisfactory and stable at a high performance.

### 4.6. Ablation studies

We carried out the ablation studies to investigate the role of the contrastive learning, i.e., $L_{cl}$, and self-attention with auxiliary classification loss for these classification tasks using incomplete multi-omics data, i.e., $L_{al}$. It should be pointed that the cross-omics autoencoders are required to handle the classification with incomplete multi-omics data, we fixed the $\lambda_{co} = 0.1$ and performed the ablation studies using the ROSMAP dataset with three omics layers. The performance comparisons of using different components among different missing rates are shown in Figure 5.

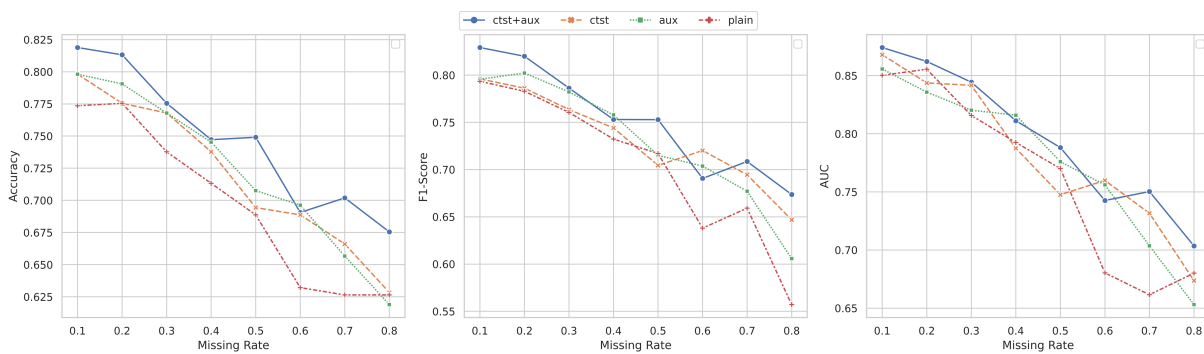

**Figure 5**. Ablation study on the ROSMAP dataset, where 'ctst' indicates that the contrastive learning loss was used and 'aux' indicates that the self-attention and auxiliary classifier were used. 'ctst+aux' indicates that both of the components were used and 'plain' indicate that neither of them was used.

By comparing the performance of the proposed model with baseline models, i.e. ctst, aux and plain models, we observed that the all components of CLCLSA promoted the model performance in most of the settings. Without employing contrastive learning, self-attention and the auxiliary classifier, the plain models achieved the lowest ACCs among different missing rates. When using either contrastive loss or auxiliary classification loss alone, the model showed a slight improvement in terms of F1-scores. However, employing both contrastive learning and auxiliary classifiers with confidence loss, the performance was significantly boosted to a higher level among different missing rates. This ablation study indicated that all loss terms play indispensable roles in these incomplete multi-omics classification tasks.

### 5. Conclusion

In this paper, we proposed a novel algorithm for multi-omics integration and classification, which can jointly exploit all training samples and is flexible for training samples with arbitrary missing omics data. Our CLCLSA model employs cross-omics autoencoders to predict the representation of missing omics data and uses contrastive learning and self-attention modules to boost model performance. Extensive experiments were conducted on four public multi-omics datasets and the experimental results indicated that the proposed CLCLSA outperformed the state-of-the-art approaches for multi-omics data classification using incomplete multi-omics data.

**Acknowledgments**

This research was supported in part by grants from the National Institutes of Health, USA (P20GM109036, R01AR069055, U19AG055373). It was also supported in part by a seed grant from Michigan Technological University Institute of Computing and Cybersystems, a graduate fellowship from Michigan Technological University Health Research Institute.